\documentclass[letterpaper, 10 pt, journal, twoside]{IEEEtran}
\ifCLASSINFOpdf
   \usepackage[pdftex]{graphicx}
   \graphicspath{{./graphic_files/}}
   \DeclareGraphicsExtensions{.pdf}
\else
\fi
\usepackage{amsmath}
\usepackage{amssymb}  
\usepackage{arydshln} 
\usepackage{multirow} 

\begin{document}
\bstctlcite{IEEEexample:BSTcontrol}
%
\title{Multi-frame Feature Aggregation for Real-time Instrument Segmentation in Endoscopic Video}
%
%
%

\author{Shan Lin$^{1}$, Fangbo Qin$^{2}$, Haonan Peng$^{1}$, Randall A. Bly$^{1}$, Kris S. Moe$^{1}$, and Blake Hannaford$^{1}$,~\IEEEmembership{Fellow,~IEEE}%
\thanks{Manuscript received: February 3, 2021; Revised May 28, 2021; Accepted June 28, 2021.}
\thanks{This paper was recommended for publication by Editor Eric Marchand upon evaluation of the Associate Editor and Reviewers' comments.} 
\thanks{$^{1}$Shan Lin, Haonan Peng, and Blake Hannaford are with Department of Electrical \& Computer Engineering, University of Washington (UW), Seattle, WA 98195, USA. Randall A. Bly and Kris S. Moe are with Department of Otolaryngology-Head \& Neck Surgery, UW, Seattle, USA.
        {\tt\footnotesize shanl3@uw.edu}}%
\thanks{$^{2}$Fangbo Qin is with Research Center of Precision Sensing and Control, Institute of Automation, Chinese Academy of Sciences, Beijing 100190, China.
        {\tt\footnotesize qinfangbo2013@ia.ac.cn}}%
\thanks{Digital Object Identifier (DOI): see top of this page.}
}

%
%

\markboth{IEEE Robotics and Automation Letters. Preprint Version. Accepted June, 2021}
{Lin \MakeLowercase{\textit{et al.}}: Multi-frame Feature Aggregation for Instrument Segmentation} 

%



\maketitle

\begin{abstract}
Deep learning-based methods have achieved promising results on surgical instrument segmentation. However, the high computation cost may limit the application of deep models to time-sensitive tasks such as online surgical video analysis for robotic-assisted surgery. Moreover, current methods may still suffer from challenging conditions in surgical images such as various lighting conditions and the presence of blood. 
We propose a novel Multi-frame Feature Aggregation (MFFA) module to aggregate video frame features temporally and spatially in a recurrent mode. 
By distributing the computation load of deep feature extraction over sequential frames, we can use a lightweight encoder to reduce the computation costs at each time step. Moreover, public surgical videos usually are not labeled frame by frame, so we develop a method that can randomly synthesize a surgical frame sequence from a single labeled frame to assist network training. We demonstrate that our approach achieves superior performance to corresponding deeper segmentation models on two public surgery datasets.
\end{abstract}

\begin{IEEEkeywords}
Computer vision for medical robotics, deep learning for visual perception, object detection, segmentation and categorization.
\end{IEEEkeywords}

%
\IEEEpeerreviewmaketitle

\section{Introduction}
%
%
%
%
\IEEEPARstart{S}{urgical} videos provide information that surgeons rely on during operations. As video-assisted and robotic-assisted surgery become more integrated, analyzing surgical videos becomes more essential. Surgical video instrument segmentation is one important task that can provide instrument locations to robotic and computer-assisted surgery systems \cite{bouget2017vision,allan20202018,shvets2018automatic}. However, challenging conditions in surgical videos such as various lighting scenarios and blood make instrument segmentation a hard problem. 
For some challenging data, identifying the instrument regions without looking at adjacent frames can be hard even for human eyes \cite{lin2020lc}, so algorithms that could leverage the temporal information for segmentation are attractive.
Although Convolutional Neural Networks (CNN) have been successfully applied in instrument segmentation \cite{shvets2018automatic,garcia2017toolnet,allan20192017,islam2019learning}, existing works usually perform segmentation on a single frame without considering the temporal information.

Recurrent Neural Networks (RNNs) are typical models for temporal sequences and have been successfully applied in many natural language processing problems \cite{chen2017enhanced,liu2019bidirectional}. 
%
In contrast, the complex gate mechanisms and the difficulty of training limit the use of RNNs in image segmentation \cite{lipton2015critical,yu2017spatio}. 
Also, RNNs usually require long temporal sequences for training and lead to a high computation burden. 

Aggregating features from neighboring frames using convolution or spatial-temporal attention-based methods
is one alternative strategy \cite{hu2020temporally}. 
In this work, we develop a Multi-frame Feature Aggregation (MFFA) module that consists of a Temporal Aggregation Block (TAB) and a Spatial Aggregation Block (SAB). TAB is designed to aggregate features of the current frame with the features of the previous frame in a sequence. SAB is then used to further aggregate non-local features based on the semantic relationships between different positions in the feature maps. 
By passing and aggregating features through neighboring frames that usually have a similar appearance, MFFA allows using lightweight encoders to reduce the computation cost at each time step.

Moreover, public surgical videos usually are not labeled by frame \cite{ross2021comparative, lin2020lc}, so a sequence of neighboring frames usually has only one frame labeled. The sparsity of labeling may make the network hard to train. To accelerate network training, we propose a simple but novel data augmentation method that can synthesize a densely labeled frame sequence by randomly moving the instrument image region of a single labeled real frame to generate synthetic frames.

This paper is among the first few studies that use temporal information for instrument segmentation. 
We evaluate our method on an endoscopic sinus surgery dataset \cite{lin2020lc} and a laparoscopic proctocolectomy dataset \cite{ross2021comparative}.
The experiment results demonstrate that by combining the proposed feature aggregation module MFFA with an existing segmentation model, we achieve promising segmentation performance with low computation costs.

\section{Related Work}

\textbf{Surgical Instrument Segmentation.} Vision-based surgical instrument segmentation aims to divide a surgical image into instrument and background regions. 
Deep CNNs have achieved promising instrument segmentation results on single images \cite{garcia2017toolnet, shvets2018automatic, islam2019learning, allan20192017, ross2021comparative}, but their performance still suffers from challenging conditions in surgical images such as various lighting scenarios, blood, and smoke.
Recently, methods that enhance feature representation by leveraging the dependencies between pixels or feature channels have been explored. 
Attia \emph{et al.} developed a RNN that captures the spatial dependencies between neighboring pixels to refine instrument segmentation \cite{attia2017surgical}. 
Ni \emph{et al.} used relationships between feature channels to emphasize instrument regions for better segmentation \cite{ni2019attention}.
Additionally, Jin \emph{et al.} proposed to propagate the instrument region detected in the previous frame using motion flow as prior information to assist instrument segmentation \cite{jin2019incorporating}.
%
Although initial works have shown promising results, studies that leverage temporal and spatial information for instrument segmentation are still limited.

\textbf{General Video Analysis Methods.}
As typical networks that can model temporal behavior, RNNs have achieved state-of-the-art performance in many video analysis tasks such as video captioning, summarization, deblurring \cite{zhao2019cam,zhao2018hsa,nah2019recurrent}. In contrast, applications of RNNs in video semantic segmentation are limited due to difficulties in training RNNs. 
In addition, 3D CNNs have been applied for video analysis, including gesture recognition and video classification \cite{zhang2020gesture, funke2019using, funke2019video}. Although achieving promising performance with their rich spatiotemporal feature representation, 3D CNNs usually have more parameters and higher computation costs than 2D CNNs.
Aggregating neighboring frame features has been proposed as an alternative approach to leverage temporal information for segmentation \cite{hu2020temporally}.
Another approach is to propagate the segmentation of the previous frame to assist segmentation on the current frame \cite{voigtlaender2019feelvos,miao2020memory}.

\section{Methods} \label{ch-label}

%
We develop a Multi-frame Feature Aggregation (MFFA) module that aggregates features both temporally and spatially for segmentation. MFFA is designed to be flexibly combined with general encoder-decoder segmentation models as introduced in Section \ref{method-overall}. 
The architecture of MFFA is presented in Section \ref{method-fam}. 

In many public datasets for surgical instrument segmentation \cite{ross2021comparative, lin2020lc}, frames are extracted from videos with a certain sub-sampling rate for labeling, so a frame sequence from a surgical video usually has only one frame labeled. The sparsity of labeling may reduce the model convergence speed, especially in the early phase of training. To compensate for the lack of densely labeled real frame sequences, we perform an initial investigation on using synthetic sequences with every frame labeled for training, as described in Section \ref{method-syn}. Each synthetic sequence consists of a real labeled frame and several frames synthesized based on this real frame.

\subsection{Task Description and Overall Framework}\label{method-overall}

Given an input sequence with $N$ video frames $X=\{x_1, x_2, ..., x_N\}$, where $x_i$ is the $i$th frame in the sequence. 
The task is to predict the corresponding binary masks $Y=\{\tilde{y}_1, \tilde{y}_2, ..., \tilde{y}_N\}$, where $\tilde{y}_i$ indicates the predicted instrument and background regions of $x_i$. For some $x_i$, the corresponding ground truth label $y_i$ is available for training or testing.

\begin{figure}[!t]
\centering
\vspace{0.5em}
\includegraphics[width=0.35\textwidth]{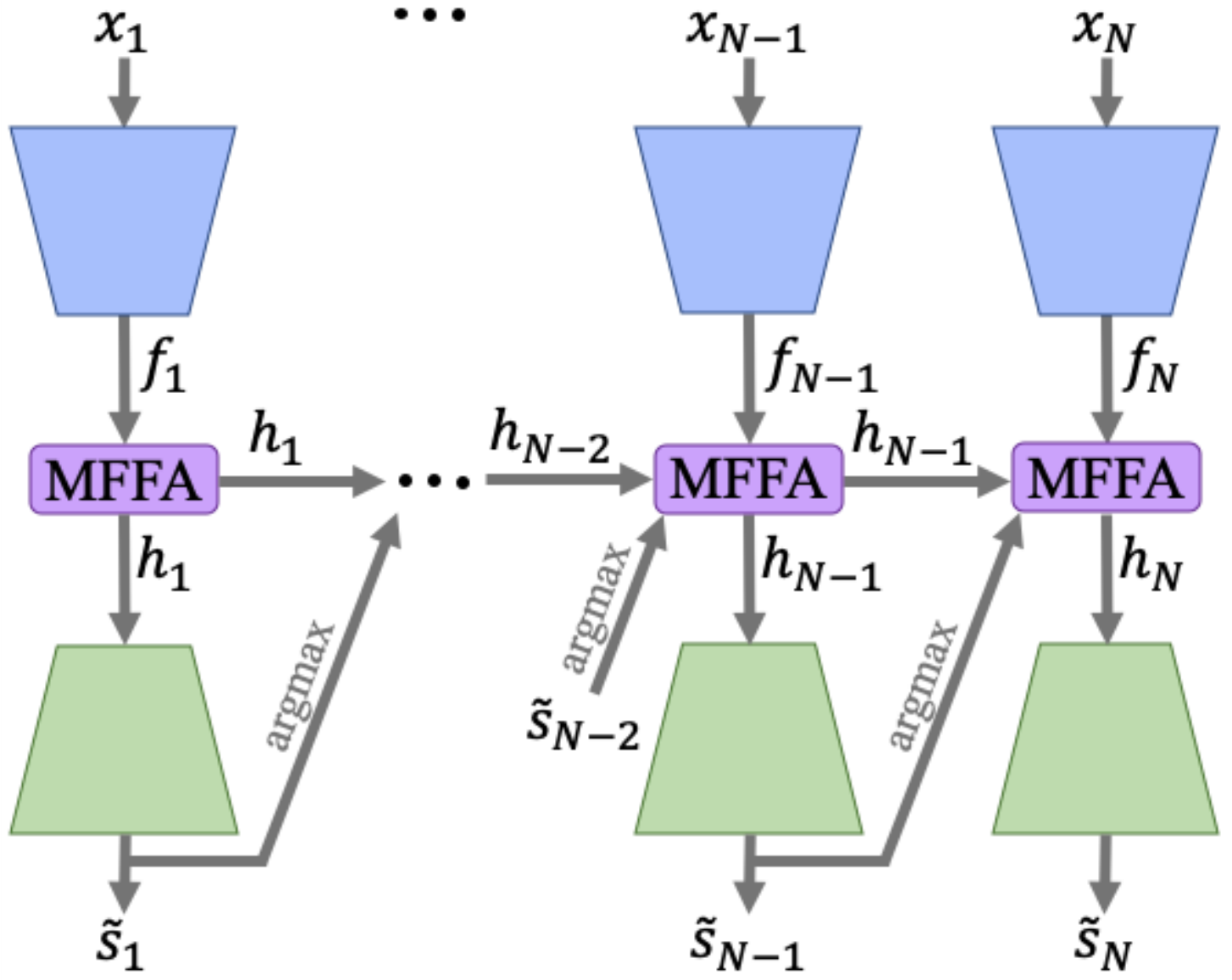}
\vspace{-1.em}
\caption{Schematic of combing MFFA with a general encoder-decoder segmentation model. MFFA is embedded between the encoder (blue) and the decoder (green). The encoder-MFFA-decoder model works in a recurrent mode and this figure illustrates the unrolled structure over time steps.} 
\vspace{-1.5em}
\label{workflow}
\end{figure}

\begin{figure*}[!t]
\centering
\vspace{0.3em}
\includegraphics[width=0.7\textwidth]{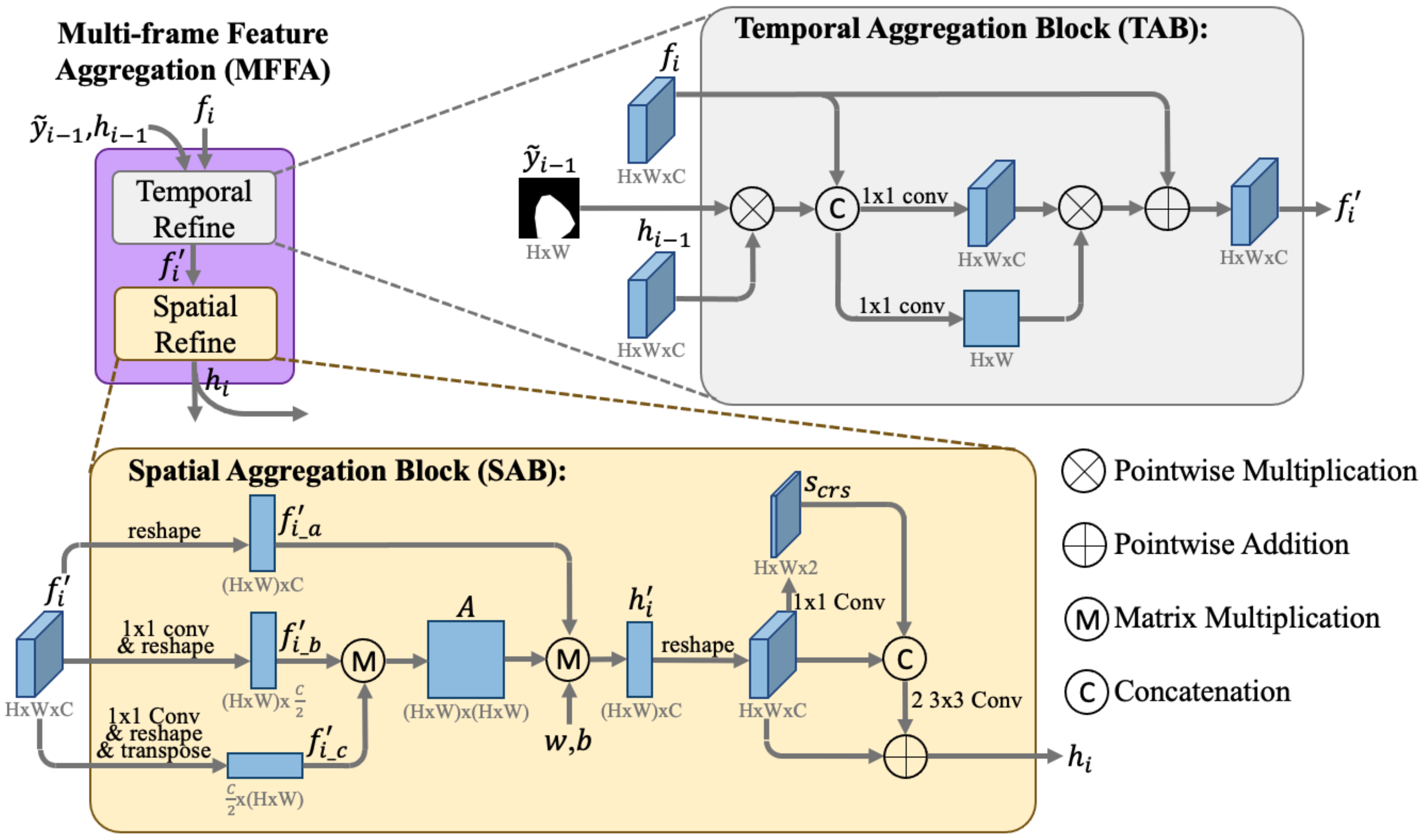}
\vspace{-1.em}
\caption{Schematic of MFFA. MFFA (purple) consists of a Temporal Aggregation Block (TAB) and a Spatial Aggregation Block (SAB).} 
\vspace{-1.5em}
\label{fam}
\end{figure*}

The proposed framework is shown in Fig. \ref{workflow}. Instead of segmenting each frame separately, we propose MFFA to aggregate features from the previous frame and pass the aggregated features to the next frame. 
MFFA is designed to combine with general segmentation models that have an encoder-decoder architecture by inserting MFFA between the encoder and the decoder.
After obtaining the feature maps $f_i$ outputted from the encoder, $f_i$, the previous output of MFFA $h_{i-1}$, and the previous predicted segmentation mask $\tilde{y}_{i-1}$ are inputted into MFFA to generate aggregated feature maps $h_i$. $h_i$ is then passed to the next frame and inputted to the decoder for segmenting the current frame. 
Finally, the encoder generates a softmax output $\tilde{s}_i$. 
The segmentation mask $\tilde{y}_i$ is then calculated with argmax from $\tilde{s}_i$.

Since MFFA is designed for general segmentation models, we demonstrate the proposed method based on a popular segmentation model DeepLabV3+ \cite{chen2018encoder} with two representative backbone feature extractors, ResNet50 \cite{he2016deep} and MobileNet \cite{howard2017mobilenets}. DeepLabV3+ and the two backbones have achieved state-of-the-art performance in many recent works on instrument segmentation \cite{ross2021comparative, ni2019attention, qin2020towards, allan20192017, sun2021lightweight}.

Moreover, because the feature maps are propagated and aggregated through frame sequences that usually consist of frames with a similar appearance, the proposed method could be considered as iterative feature aggregation. Therefore, we propose to use lightweight encoders to extract $f_i$. 
%
More specifically, we use ResNet50 from the beginning to block3 as its trimmed version, named ResNet50-b3 \cite{he2016deep}. We use MobileNet from the beginning to the $8$th pointwise convolution as its trimmed version named MobileNet-p8 \cite{howard2017mobilenets}. 

\subsection{Objective Functions} \label{model-loss}
To train the proposed model, we extract real frame sequences from the surgical videos. However, as discussed at the beginning of Section \ref{ch-label}, sparsely labeled real frame sequences might influence model training efficiency. Therefore, we propose a method that can generate synthetic frame sequences with every frame labeled (see Section \ref{method-syn}) to explore if the model can benefit from training with densely labeled frame sequences. Specifically, we compare the models obtained under these two settings: i) Train the model with only the real frame sequences; ii) Train the model with only the synthetic frame sequences in the first half of the training phase, and further train the model with only the real frame sequences in the second half of the training phase.

For training, we calculate the cross-entropy loss to evaluate the segmentation performance for every $\tilde{s}_i$ that has ground truth $y_i$. After getting the one-hot encoding for $y_i$ as $s_i$, the cross-entropy loss can be calculated as 
\begin{equation}
\begin{aligned}
\mathcal{L}_{CE}(s_i,\tilde{s}_i) &= -\frac{1}{M}\sum_k(s_i)_k\log(\tilde{s}_i)_k
\end{aligned}
\end{equation}
where $M$ is the number of elements in the segmentation map, $(s_i)_k$ is the $k$-th element of $s_i$ and $(\tilde{s}_i)_k$ is the $k$-th element of $\tilde{s}_i$.

When the network is trained with synthetic frame sequences, we input the sequence both forward and backward to the network and calculate the average cross-entropy loss of all frames.
The first frame in a sequence does not have features passed from a previous frame, so only spatial feature aggregation can be performed on it (see Section \ref{method-fam}). Therefore, the segmentation results and the losses will be different when passing a sequence forward and backward. To indicate whether a segmentation map is predicted with or without the information from a previous frame, we change the symbol of the predicted softmax output from $\tilde{s}_i$ to $\tilde{s}_i^j$. For a sequence of $N$ frames, a $j\in[1,N]$ means the aggregated features of the $j$-th frame are used; otherwise, the segmentation is obtained without temporal information. Then the forward and backward average cross-entropy losses are given by
\begin{equation}
\begin{aligned}
\mathcal{L}_{fw} = \frac{1}{N}\sum_{i=1}^N\mathcal{L}_{CE}(s_i,\tilde{s}_i^{i-1})
\end{aligned}
\end{equation}
\begin{equation}
\begin{aligned}
\mathcal{L}_{bw} = \frac{1}{N} \sum_{i=N}^1\mathcal{L}_{CE}(s_i,\tilde{s}_i^{i+1})
\end{aligned}
\end{equation}
When the network is trained with real frame sequences that have only the last frame labeled, we pass the sequence forward and evaluate the segmentation accuracy on the last frame. We use $\mathcal{L}_{last}$ to represent this loss and it is given by
\begin{equation}
\begin{aligned}
\mathcal{L}_{last} = \mathcal{L}_{CE}(s_N,\tilde{s}_N^{N-1})
\end{aligned}
\end{equation}
For each synthetic or real sequence, there is only on labeled
real frame. To guarantee the network could extract good enough features from the first frame in the sequence which does not have features propagated from a previous frame, we input the labeled real frame and calculate its cross-entropy loss $\mathcal{L}_{1st}$. Assume $x_k$ is the real labeled frame in a sequence, then we have
\begin{equation}
\begin{aligned}
\mathcal{L}_{1st} = \mathcal{L}_{CE}(s_k,\tilde{s}_k^0)
\end{aligned}
\end{equation}

Finally, when only the synthetic sequences are used for training, the overall objective function is given by 
\begin{equation}
\begin{aligned}
\mathcal{L}&= \lambda_1\mathcal{L}_{fw}+\lambda_2\mathcal{L}_{bw}+\lambda_3\mathcal{L}_{1st}
\end{aligned}
\end{equation}
When only the real sequences are used for training, the overall objective function is given by
\begin{equation}
\begin{aligned}
\mathcal{L}&= \lambda_4\mathcal{L}_{last}+\lambda_5\mathcal{L}_{1st}
\end{aligned}
\end{equation}
where $\lambda_i$ are hyper-parameters that balance the impact of the losses. We choose $\lambda_1=\lambda_2=\lambda_3=\frac{1}{3}$ and $\lambda_4=\lambda_5=\frac{1}{2}$.

\begin{figure}[!t]
\centering
\includegraphics[width=0.36\textwidth]{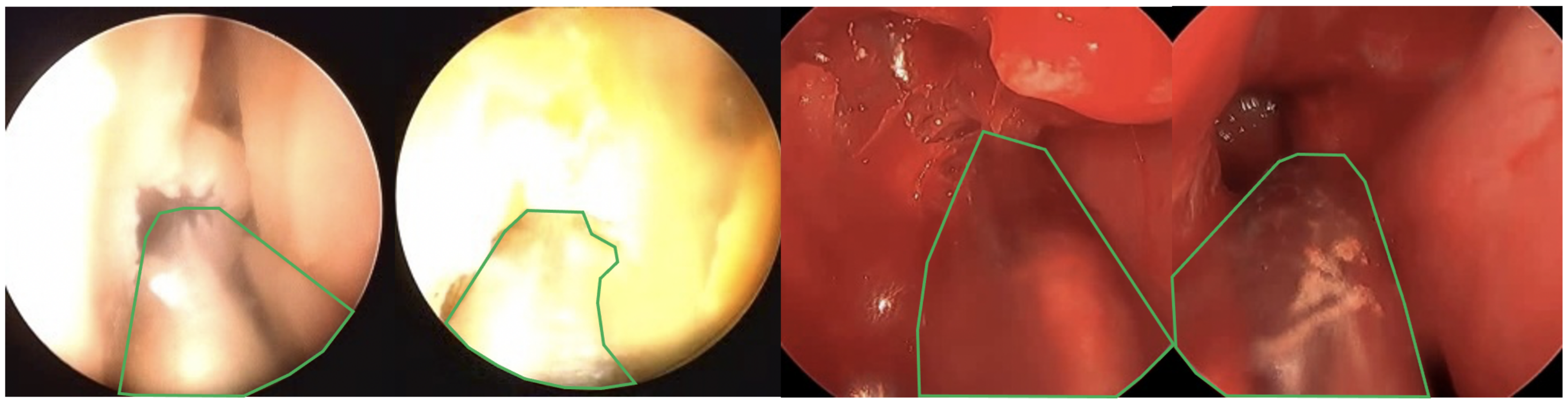}
\vspace{-1.0em}
\caption{Examples of endoscopic sinus surgery video frames with specular reflection or background reflection on the instrument surface. The instrument contours are marked with green lines.} 
\vspace{-1.5em}
\label{sample_reflection}
\end{figure}

\subsection{Multi-frame Feature Aggregation (MFFA)}\label{method-fam}

MFFA consists of a Temporal Aggregation Block (TAB) and a Spatial Aggregation Block (SAB) in series as shown in Fig. \ref{fam}. MFFA requires $\tilde{y}_{i-1}$, $h_{i-1}$ and $f_i$ as inputs and outputs $h_i$. 
In endoscopic sinus surgery, distinguishing instruments from the background can be very challenging 
due to reflection or the presence of blood as shown in Fig. \ref{sample_reflection}. 
%
Because instrument locations are usually close in neighboring video frames, we propose to emphasize the instrument regions of $f_i$ by using the instrument features of $h_{i-1}$ for aggregation in TAB. The instrument features are extracted by element-wise multiplication between the previous output $y_{i-1}$ and $h_{i-1}$. 
The extracted previous instrument features and current features are then concatenated and passed to two parallel 1x1 convolutions. One of the convolutions performs feature aggregation. The other convolution is followed by a sigmoid function. It estimates the similarity between the previous and current feature maps, and gives a 2D weight map that serves as a gate to regulate the aggregated features. 
The aggregated features are then multiplied with the corresponding weights and finally added by $f_i$ to generate $f'_i$ as the output of TAB. 
%
When applying MFFA to the first frame in a sequence, the previous output and features are not available, so we skip TAB and directly input $f_i$ to SAB.

After temporal feature aggregation in TAB, the generated features $f'_i$ is further aggregated using SAB based on spatial relationships. SAB is inspired by the spatial self-attention module \cite{fu2019dual} and Graph Attention Network (GAT) \cite{lu2020cnn}. Given $f'_i\in\mathbb{R}^{H\times W \times C}$, $f'_i$ can be considered as a graph with $H\times W$ vertices. To estimate the edge values of the graph, we use part of the spatial self-attention module to calculate a spatial attention matrix $A\in\mathbb{R}^{(H\times W)\times(H\times W)}$ \cite{fu2019dual}.
Specifically, $f'_i$ is passed to two parallel 1$\times$1 convolutions for generating two new feature maps $\{f'_{i\_b},f'_{i\_c}\}\in\mathbb{R}^{H\times W \times \frac{C}{2}}$, 
which are then reshaped to $\mathbb{R}^{(H\times W)\times \frac{C}{2}}$. Next we transpose $f'_{i\_c}$ to $\mathbb{R}^{\frac{C}{2}\times(H\times W)}$ and perform matrix multiplication between $f'_{i\_b}$ and $f'_{i\_c}$ to get the spatial attention matrix $A$. 
Each element of $A$ represents the similarity between the corresponding positions in $f'_i$.
Then we reshape $f'_i$ to $\mathbb{R}^{(H\times W)\times C}$ and aggregate the vertices features using $A$ through matrix multiplication. After that we use trainable parameters $w\in\mathbb{R}^{C\times C}$ and bias $b\in\mathbb{R}^C$ to get a refined features $h'_i$. 
The expression of this process is given by 
\begin{equation}
\begin{aligned}
h'_i &= Af'_{i\_a}w+b\\
\end{aligned}
\end{equation}
After reshaping $h'_i$ from $\mathbb{R}^{(H\times W)\times C}$ to $\mathbb{R}^{H\times W\times C}$, we further refine the features using a ResNet module-like block proposed in \cite{zhang2019canet}. Specifically, $h'_i$ goes through a 1$\times$1 convolution and the softmax function to generate a coarse segmentation $\tilde{s}_{crs}$. Finally, the output of SAB is obtained by
\begin{equation}
\begin{aligned}
h_i &= h'_i+\Phi(h'_i,\tilde{s}_{crs})\\
\end{aligned}
\end{equation}
where $\Phi(a,b)$ performs $a$ and $b$ concatenation followed by two 3$\times$3 convolution filters in series. 

\subsection{Synthetic Frame Sequence}\label{method-syn}
We propose a method that can generate a synthetic frame sequence from a real labeled frame.
The proposed sequence synthesis method is developed based on the data augmentation method introduced in \cite{fang2019instaboost}. In \cite{fang2019instaboost}, the dataset was augmented by cropping and randomly jittering the target objects in images. The holes left by the moved target objects were then filled with an off-the-shelf inpainting method \cite{telea2004image}. 

\begin{figure}[!t]
\centering
\vspace{0.3em}
\includegraphics[width=0.36\textwidth]{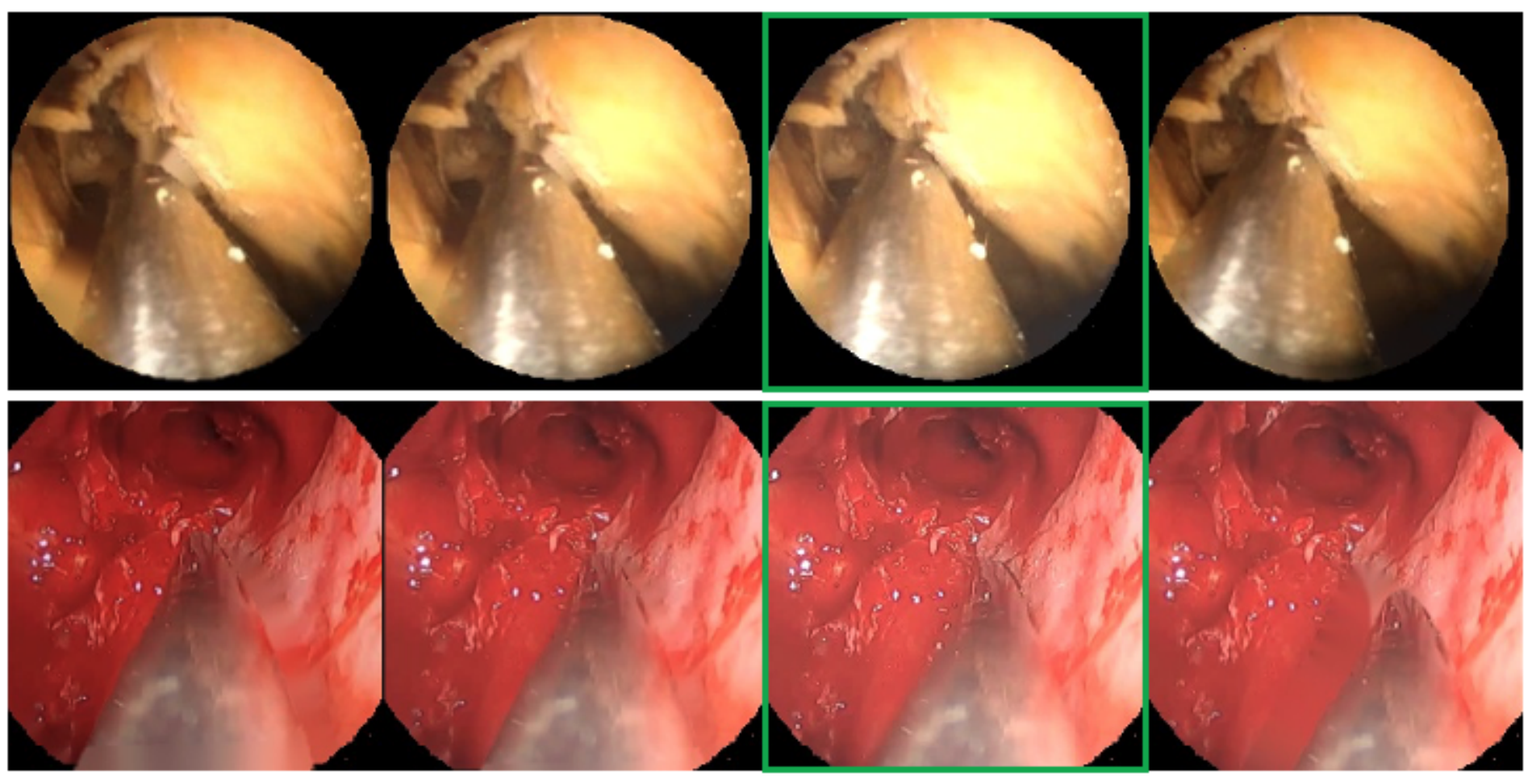}
\vspace{-1.em}
\caption{Examples of synthetic frame sequences. The top row is a sequence generated from a cadaveric surgical frame and the bottom row is a sequence generated from a live surgical frame. The real frames are bounded by green boxes and others are synthetic frames augmented from the real frames.}
\vspace{-1.5em}
\label{syn_sample}
\end{figure}

Different from \cite{fang2019instaboost}, we need frame sequences with each frame labeled for training. 
Given a real frame $x$ and its label $y$, our goal is to generate a sequence with $N$ labeled frames $Z=\{z_1, z_2, ..., z_N\}$. We first put $x$ to the center of the target sequence, i.e. let $z_C=x$ where $C=\lfloor\frac{N+1}{2}\rfloor$. 
For the instrument in the $i$th frame, we define its translations on the x and y-axis and the rotation with respect to the instrument in $x$ as $dx_i$, $dy_i$ and $d\theta_i$, which are called moving parameters. 
Then we randomly select the moving parameters for the first and the last frames (i.e., $dx_0,dy_0,d\theta_0,dx_N,dy_N,d\theta_N$) and use linear interpolation to decide the moving parameters of the other frames in the target sequence. Finally, the instruments are cropped, rotated and moved to corresponding locations using the aforementioned augmentation method to generate all synthetic frames that form the target sequence. We also apply the same process with the same moving parameters on $y$ to get the corresponding labels. 
%
Fig. \ref{syn_sample} shows examples of synthetic frame sequences.

\section{Datasets}\label{dataset}

We evaluate our model on two public datasets of different surgery types: endoscopic sinus surgery dataset UW-Sinus-Surgery-C/L \cite{lin2020lc} and laparoscopic proctocolectomy dataset ROBUST-MIS-Proctocolectomy \cite{ross2021comparative, maier2021heidelberg}. 

\textbf{UW-Sinus-Surgery-C/L} 
consists of a cadaver and a live endoscopic sinus surgery dataset \cite{lin2020lc}. The cadaver dataset consists of 10, 5-23 minute cadaveric surgery videos with a resolution of 320$\times$240. A total of 4345 frames were extracted with a sampling rate of 0.5 Hz from the videos. The instrument regions were manually labeled for training semantic segmentation. The live dataset consists of 3, 12-66 minute live surgery videos with a resolution of 1920$\times$1080. Similarly, a total of 4658 frames were extracted with a sampling rate of 1 Hz and labeled. 

The main difficulties of this dataset include background reflection on the instrument surface, specular reflection, blur from motion, blood and smoke. 
In both cadaveric and live surgery videos, reflections make the instruments have similar appearances to the background. In the live surgery videos, segmentation becomes even harder with instruments colored by blood. The instruments are usually used at low speeds during surgery, but senior surgeons who are more familiar with the anatomies may move instruments faster, especially when switching surgery sites. Also, smoke generated when electrocauteries are used can make the instruments hard to identify. Considering that these challenging conditions usually are not consistent through the entire video but switch between each other, temporal information could be useful to improve segmentation performance.

Endoscopic frames have black border regions that contain no useful information, so we extracted frame center regions and downscaled them to 240$\times$240 for segmentation. To evaluate the generalization ability of our model, we performed K-fold cross-validation with $K=3$. We use the same fold split rules as stated in \cite{lin2020lc}. Specifically, the 10 videos in the cadaver dataset are split into the following 3 folds: 4 videos (Procedure ID: 1, 2, 3, and 4), 3 videos (ID: 5, 6, and 7), and 3 videos (ID: 8, 9, and 10). For the live dataset, the 3 folds are formed by the 1\textsuperscript{st}, 2\textsuperscript{nd}, and the 3\textsuperscript{rd} videos.

\textbf{ROBUST-MIS} was published for MICCAI 2019 Robust Medical Instrument Segmentation challenge, and it consists of a proctocolectomy, a rectal resection, and a sigmoid resection dataset \cite{ross2021comparative, maier2021heidelberg}. As an initial exploration, we evaluated our method only on the proctocolectomy dataset for the binary segmentation task. The proctocolectomy dataset has a training set of 2943 960$\times$540 labeled frames sampled from 8, 3-5 hour laparoscopic videos corresponding to 8 procedures. 
The difficulties of this dataset include reflection, blood, smoke, various lighting conditions, small and transparent instruments, and blur from motion. 
Similarly, we performed 3-fold cross-validation to evaluate our model. Based on the number of frames from each procedure, we split the given training data into the following 3 folds: 6 videos (ID: 1, 2, 3, 4, 5, and 8), 1 video (ID: 9), and 1 video (ID: 10).

\section{Implementation Details}\label{details}
The segmentation models were implemented on a 4.20GHz Intel i7-7700K CPU and a Nvidia Titan Xp GPU.

\begin{table*}[!t]
\caption{\vspace{-0.3em} Segmentation Performance on UW-Sinus-Surgery-C/L}
\label{Tab_seg}
\vspace{-1.em}
\centering
\begin{tabular}{c|c|c:c|c|c|c}
\hline
\multirow{2}*{Group}&\multirow{2}*{Model(Backbone)} & \multicolumn{2}{c|}{Proposed} & \multicolumn{2}{c|}{Performance (mDSC($\%$) / mIoU($\%$))} & Time\\ \cline{3-6}
& & MFFA & Synth. & Sinus-Surgery-C & Sinus-Surgery-L & (ms)\\
\hline
\hline
\multirow{3}*{1}&TernausNet-16 \cite{iglovikov2018ternausnet} (VGG16) & \multirow{3}*{n/a} & \multirow{3}*{n/a} & 85.4(2.2)/80.1(2.7) & 79.5(5.9)/73.4(6.8) & 12.9\\
&LWANet \cite{ni2019attention} (MobileNet) & & & 81.1(3.8)/74.9(4.5) & 72.6(7.7)/65.2(8.8) & 5.3\\
&MAFA-DL3+ \cite{qin2020towards} (ResNet50) & & & 91.2(1.0)/86.8(1.2) & 87.7(3.8)/82.1(4.6) & 19.3\\
\hline
\hline
\multirow{4}*{2}&DL3+ \cite{chen2018encoder} (MobileNet) & $\times$ & $\times$ & 81.4(4.2)/75.5(4.8) & 76.6(6.8)/69.5(8.0) & 3.1\\
\cdashline{2-2}
&& $\times$ & $\times$ & 80.4(3.5)/74.0(4.4) & 75.7(7.0)/68.0(8.3) & 2.1\\
&DL3+ \cite{chen2018encoder} (MobileNet-p8) & \checkmark & $\times$ & 84.8(2.9)/79.4(3.3) & 81.3(6.0)/74.7(7.1) & 2.9\\
&& \checkmark & \checkmark & \textbf{86.4(2.2)/81.1(2.7)} & \textbf{83.2(4.1)/77.0(5.0)} & 2.9\\
\hline
\multirow{4}*{3}&DL3+ \cite{chen2018encoder} (ResNet50) & $\times$ & $\times$ & 86.0(2.0)/81.0(2.5) & 81.8(5.4)/75.2(6.4) & 8.2\\
\cdashline{2-2}
&& $\times$ & $\times$ & 83.9(5.0)/78.8(5.6) & 80.1(5.4)/73.4(6.4) & 5.0\\
&DL3+ \cite{chen2018encoder} (ResNet50-b3) & \checkmark & $\times$ & 86.1(2.5)/81.0(3.0) & 83.2(4.9)/77.0(5.7) & 5.4\\
&& \checkmark & \checkmark & \textbf{88.9(1.4)/84.0(1.8)} & \textbf{85.8(3.3)/80.0(4.3)} & 5.5\\
\hline
\multicolumn{7}{l}{* i) The mDice and mIoU are shown in `mean(standard deviation)'; ii) MobileNet-p8 is the trimmed MobileNet}\\
\multicolumn{7}{l}{$~~$and ResNet50-b3 is the trimmed ResNet50; iii) `Synth.' indicates that the synthetic data are used in the first 20}\\
\multicolumn{7}{l}{$~~$training epochs; v) The bold font indicates the best performance in the column of each group.}\\
\end{tabular}
\vspace{-2.em}
\end{table*}

\begin{table}[!t]
\caption{\vspace{-0.3em} Conover Post Hoc Test Results of Baselines (DeepLabV3+ \cite{chen2018encoder}) and Proposed Models on UW-Sinus-Surgery-C/L Dataset}
\label{Tab_ttest}
\vspace{-1.em}
\centering
\begin{tabular}{c:c|c}
\hline
\multicolumn{2}{c|}{Compared Model (Backbone,\textbf{R}eal/\textbf{S}ynthetic Data)} & p-value\\ \cline{2-3} 
\hline
& DL3+(MobileNet-p8,R) & 0.83\\
DL3+(MobileNet,R) & MFFA-DL3+(MobileNet-p8,R) & 0.08\\
& MFFA-DL3+(MobileNet-p8,S) & ~0.02*\\
\hline
\multirow{4}*{DL3+(ResNet50,R)} & DL3+(ResNet50-b3,R) & 0.38\\
& MFFA-DL3+(ResNet50-b3,R) & 0.74\\
& MFFA-DL3+(ResNet50-b3,S) & 0.07\\
& MFFA-DL3+(MobileNet-p8,S) & 0.83\\
\hline
MFFA-DL3+ & MFFA-DL3+ & \multirow{2}*{0.51}\\
(MobileNet-p8,R) & (MobileNet-p8,S) & \\
\hline
MFFA-DL3+ & MFFA-DL3+ & \multirow{2}*{0.13}\\
(ResNet50-b3,R) & (ResNet50-b3,S) & \\
\hline
\multicolumn{3}{l}{$~~$* i) In the brackets, `R' represents only real data are used for}\\
\multicolumn{3}{l}{$~~$training, while `S' represents both synthetic and real data are}\\
\multicolumn{3}{l}{$~~$used for training; ii) p-values $<0.05$ are marked with `*'.}\\
\end{tabular}
\vspace{-2.0em}
\end{table}

\textbf{Instrument segmentation models.}
The segmentation models were implemented with or without the proposed MFFA module. The backbones of all models were pre-trained on ImageNet. MFFA was implemented based on Fig. \ref{fam} by setting $C=128$ and adding a ReLU activation function at the output of TAB and after all convolutions in SAB. For UW-Sinus-Surgery-C/L, we used the Adam optimizer \cite{kingma2014adam} to train each model with 40 epochs and 16 batch size. The learning rate was initialized as 0.0005 and exponentially decayed every 5 epochs from the 20th epoch with a decay rate of 0.5. To evaluate the models on ROBUST-MIS-proctocolectomy, we chose a batch size of 8 while keeping the epoch and learning rate settings the same.

When we trained the models without MFFA, we augmented images by i) changing hue, brightness, saturation and contrast; ii) flipping, rotation, scaling, and cropping. When using UW-Sinus-Surgery-C/L, we cropped the frames to a resolution of 192$\times$192 to accelerate training, while the full images with a resolution of 240$\times$240 were used for testing. For ROBUST-MIS-proctocolectomy, the raw images were resized to 640$\times$360 and kept in the same size after augmentation; during testing, the raw images were resized to 640$\times$360 for inference, and then the output segmentation maps were resized back to 960$\times$540 to evaluate the performance. When we trained the models with MFFA, we implemented similar data augmentation on the frame sequences. We jittered the appearance of every frame in the sequences independently, while all frames in the same sequence shared the same parameters for flipping, rotation, scaling, and cropping.

Additionally, on UW-Sinus-Surgery-C/L, we compared the models (with MFFA) that were either: i) trained with only the real frame sequences or ii) trained with only the synthetic frame sequences in the first 20 epochs and trained with only the real frame sequences in the last 20 epochs. Details of these two training settings can be found in Section \ref{model-loss}. The second setting is represented with `Synth.' in Table \ref{Tab_seg} and Table \ref{Tab_ablation}. For ROBUST-MIS-proctocolectomy, we only used the real frame sequences for training.

\textbf{Synthetic and real frame sequence.}
Considering the computation ability, we train models with sequences of 4 frames. To generate synthetic sequences, we need to decide the translations and rotation angles of the instruments in the first and the last frames with respect to the instrument in the labeled frame. The translation values were randomly selected from a uniform distribution over 15 to 40 pixels on both positive/negative x-axis and y-axis directions. The rotation angles were randomly selected from a uniform distribution over -30 to 30 degrees. Moreover, real frame sequences were extracted around labeled frames in the videos and each frame in a sequence was sampled out of 3 consecutive frames. 


\section{Experiments and Results}

We combined MFFA with DeepLabV3+ (DL3+) \cite{chen2018encoder} (abbreviated as MFFA-DL3+) and compared with the original version of DeepLabV3+ \cite{chen2018encoder} as baseline. MFFA-DL3+ were also compared with advanced segmentation models include TernausNet \cite{iglovikov2018ternausnet}, LWANet \cite{ni2019attention} and MAFA \cite{qin2020towards}.

\subsection{Evaluation}
For evaluating the segmentation performance, we used Dice similarity coefficient (DSC) and Intersection over Union (IoU) \cite{taha2015metrics}, which are defined as
$$
DSC = \frac{2|X\cap Y|}{|X|+|Y|},IoU = \frac{|X\cap Y|}{|X\cup Y|}
$$
where $X$ and $Y$ are the predicted and ground truth segmentation maps, respectively. Further, we estimate the computation cost of each model using its inference time, which is the time for the model to predict a segmentation map.

The models were evaluated with 3-fold cross-validation as described in Section \ref{dataset} and a video can only be used either for training or testing in each of the three validations. The trained models were tested by propagating the aggregated feature maps and performing segmentation every third frame throughout the test videos, and the segmentation performance was evaluated on the labeled frames.

\subsection{Experiments on UW-Sinus-Surgery C/L}
The segmentation results on UW-Sinus-Surgery C/L is shown in Table \ref{Tab_seg}. There are three groups of methods: i) Group 1 consists of three advanced segmentation models on single frames; ii) Group 2 consists of the original version of DL3+(MobileNet), DL3+(MobileNet-p8), and MFFA-DL3+(MobileNet-p8) trained with/without synthetic sequences; iii) Group 3 consists of the original version of DL3+(ResNet50), DL3+(ResNet-b3), and MFFA-DL3+(ResNet-b3) trained with/without synthetic sequences. 
To determine if there are significant differences between the eight models in Group 2 and 3, we performed Friedman test \cite{friedman1937use} followed by a Conover post hoc test \cite{conover1979multiple}
on the 6 data folds of UW-Sinus-Surgery C/L. Friedman test shows that there is a statistically significant difference between the performance of these eight models with a p-value$<0.05$ and the pairwise comparisons calculated using Conover's test are shown in Table \ref{Tab_ttest}. In each row of Table \ref{Tab_ttest}, a p-value$<0.05$ indicates that there is a statistically significant difference between the corresponding paired comparison methods.

When both synthetic and real frame sequences are used for training, MFFA-DL3+ with reduced backbone achieved superior performance of 3.1$\%$$\sim$9.5$\%$ better mDSC and 3.3$\%$$\sim$10.7$\%$ better mIoU with less inference time compared with the baseline DL3+ \mbox{\cite{chen2018encoder}}. The third row of Table \ref{Tab_ttest} further shows that there is a statistically significant difference between MFFA-DL3+(MobileNet-p8) trained with synthetic frame sequences and DL3+(MobileNet). Also, MFFA-DL3+(MobileNet-p8) achieved comparable results to DL3+(ResNet50) with a p-value of 0.83 while used only about 35\% of the average inference time. 
Moreover, compared with the three advanced segmentation models shown in group 1, both MFFA-DL3+(MobileNet-p8) and MFFA-DL3+(ResNet-b3) achieved superior or comparable performance with much lower computation costs.

\begin{figure}[!t]
\centering
\vspace{0.2em}
\includegraphics[width=0.32\textwidth]{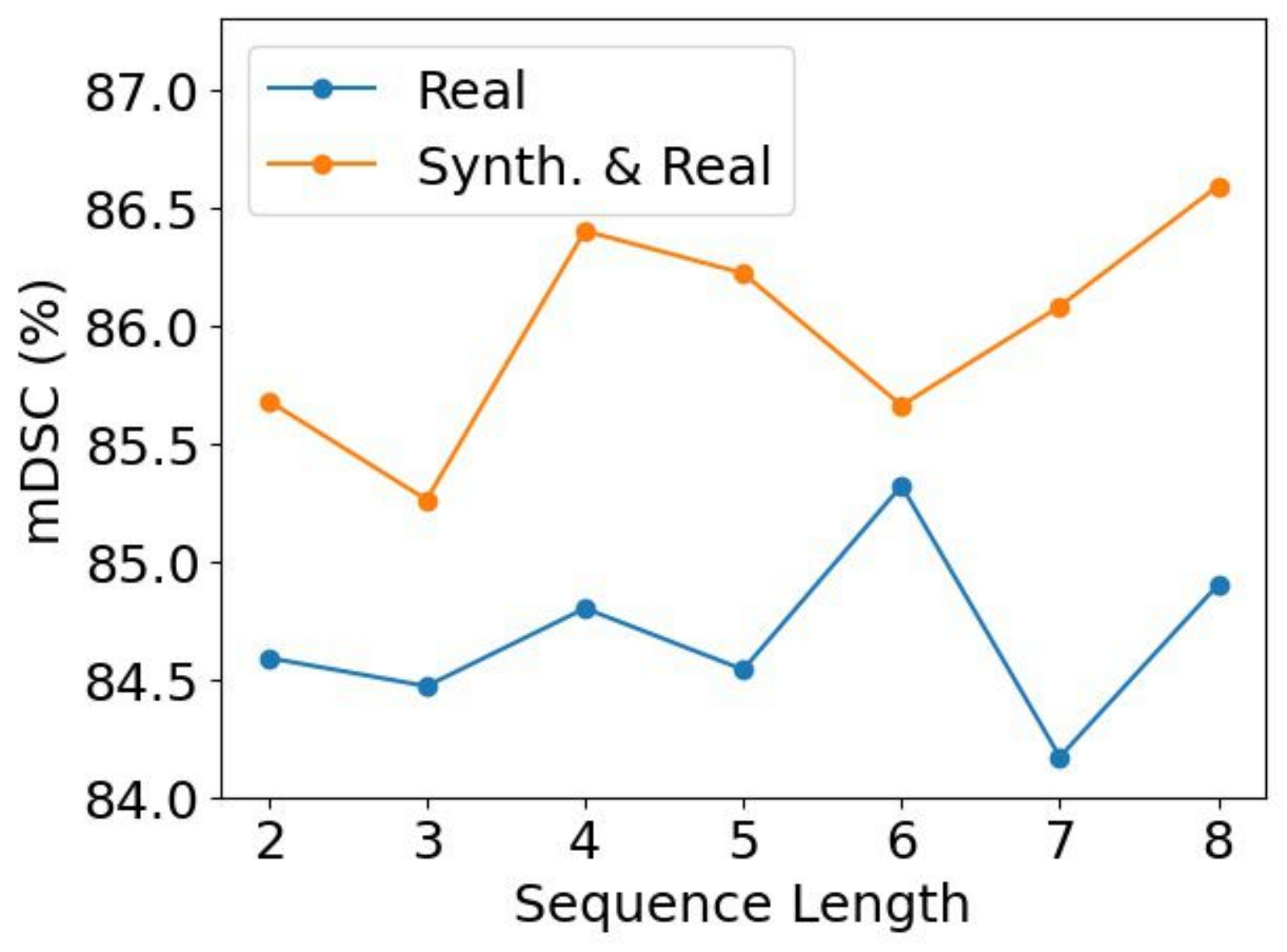}
\vspace{-1.em}
\caption{mDSC achieved by MFFA-DL3+(MobileNet-p8) on UW-Sinus-Surgery C/L with sequences of different lengths. The blue line ('Real') shows the performance of models trained only with real sequences. The yellow line ('Synth. \& Real') shows the performance of models trained with both synthetic and real sequences.}
\vspace{-1.5em}
\label{seq-length-sensitivity}
\end{figure}

\begin{figure*}[!t]
\centering
\vspace{0.5em}
\includegraphics[width=0.99\textwidth]{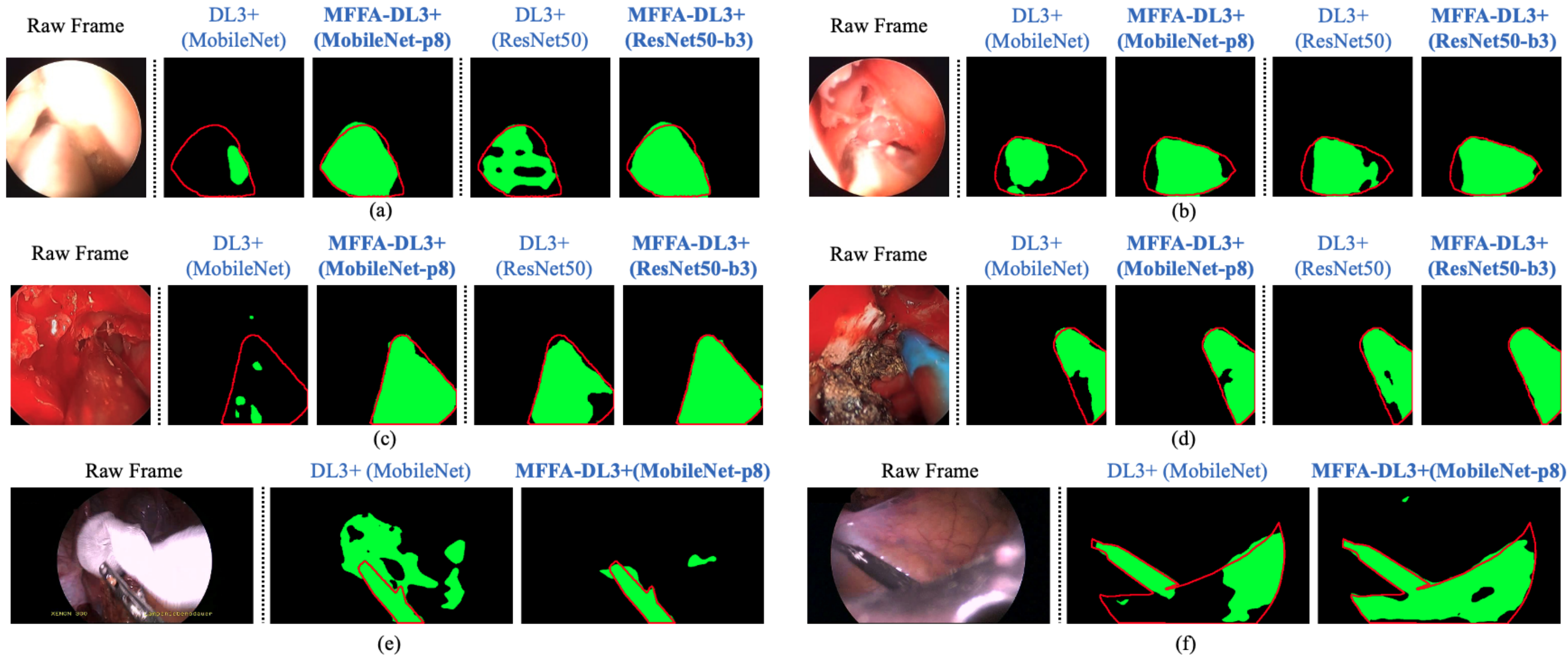}
\vspace{-1.0em}
\caption{Examples of segmentation results. In (a-d), the first frame is the input raw frame and the next four frames are the segmentation results of DeepLabV3+ (DL3+) \cite{chen2018encoder} with MobileNet as the backbone feature extractor, DL3+ with MobileNet-p8 and MFFA, DL3+ with ResNet50, and DL3+ with ResNet50-b3 and MFFA, respectively. In (e-f), the first frame is the input raw frame and the next two frames are the segmentation results of DL3+ with MobileNet, and DL3+ with MobileNet-p8 and MFFA, respectively. MFFA-DL3+(MobileNet-p8) and MFFA-DL3+(ResNet50-b3) were trained with both synthetic and real frame sequences. 
The predicted instrument regions are drawn in green and the true instrument contours are labeled by red lines. 
The example results of UW-Sinus-Surgery-C, UW-Sinus-Surgery-L, ROBUST-MIS-Proctocolectomy are shown in (a,b), (c,d), and (e,f), respectively. }
\vspace{-1.5em}
\label{trans_rst_ex}
\end{figure*}

To evaluate the effectiveness of i) Temporal Aggregation Block (TAB), ii) Spatial Aggregation Block (SAB), and iii) $\mathcal{L}_{1st}$, we conducted ablation studies on UW-Sinus-Surgery C/L with DeepLabV3+ \cite{chen2018encoder}, as shown in Table \ref{Tab_ablation}. 
The original version of DL3+(MobileNet) was used in experiment 1, while DL3+ with a trimmed backbone MobileNet-p8 was used in experiment 2$\sim$8. Because TAB is responsible for aggregating temporal information, the model was trained with frame sequences only when TAB was used (experiment 4-8).
Compared with experiment 2, experiment 3 shows that using SAB without temporal feature aggregation improved the performance on both the cadaver and live datasets, while experiment 4 shows that TAB improved the segmentation performance by leveraging the temporal information. Further, experiment 5 shows that better performance was achieved by using TAB and SAB together.
By comparing experiments 5 and 6, and comparing experiments 7 and 8, we found $\mathcal{L}_{1st}$ is more effective when training with synthetic frame sequences.

\begin{table*}[!t]
\vspace{1.em}
\caption{\vspace{-0.3em} Ablation Studies of MFFA with DeepLabV3+ on UW-Sinus-Surgery-C/L Dataset}
\label{Tab_ablation}
\vspace{-1.em}
\centering
\begin{tabular}{c|c:c:c:c:c|c|c}
\hline
\multirow{2}*{No.} & \multicolumn{5}{c|}{Method} & \multicolumn{2}{c}{Performance (mDSC($\%$) / mIoU($\%$))}\\
\cline{2-8}
&Backbone & TAB & SAB & Synth.&$\mathcal{L}_{1st}$& Sinus-Surgery-C & Sinus-Surgery-L\\
\hline
1 &MobileNet&$\times$&$\times$&n/a&n/a& 81.4(4.2)/75.5(4.8) & 76.6(6.8)/69.5(8.0)\\
\cdashline{2-2}
2 &\multirow{7}*{MobileNet-p8}&$\times$&$\times$&n/a&n/a& 80.4(3.5)/74.0(4.4) & 75.7(7.0)/68.0(8.3)\\
3 &&$\times$&\checkmark&n/a&n/a& 82.4(1.7)/76.6(2.2) & 79.1(5.4)/71.7(6.6)\\
4 &&\checkmark&$\times$&$\times$&$\times$& 83.0(3.3)/76.6(4.0) & 77.4(7.0)/69.9(8.4)\\
5 &&\checkmark&\checkmark&$\times$&$\times$& 83.8(1.6)/78.2(2.0) & 80.1(5.5)/73.2(6.6)\\
6 &&\checkmark&\checkmark&$\times$&\checkmark& 84.8(2.9)/79.4(3.3) & 81.3(6.0)/74.7(7.1)\\
7 &&\checkmark&\checkmark&\checkmark&$\times$& 84.3(1.7)/78.9(2.3) & 79.8(5.3)/73.3(6.2)\\
8 &&\checkmark&\checkmark&\checkmark&\checkmark& \textbf{86.4(2.2)/81.1(2.7)} & \textbf{83.2(4.1)/77.0(5.0)}\\
\hline
\multicolumn{8}{l}{* The bold font indicates the best performance in the column.}\\
\end{tabular}
\vspace{-2.0em}
\end{table*}

Further, we assessed the sensitivity of segmentation performance to the frame sequence length on UW-Sinus-Surgery-C with MFFA-DL3+(MobileNet-p8). Fig. \ref{seq-length-sensitivity} shows the test mDSC with frame sequence length ranges from 2 to 8. We found training the model with synthetic sequences achieved better performance than when only real sequences are used. However, limited by the computation resource, the model cannot be tested on longer sequences, which are needed to draw a conclusion on the relationships between sequence length and segmentation performance.

\subsection{Experiments on ROBUST-MIS-Proctocolectomy}
We compared the performance of DL3+(MobileNet), DL3+(MobileNet-p8), and MFFA-DL3+(MobileNet-p8) on ROBUST-MIS-Proctocolectomy, as shown in Table \ref{Tab_endovis_seg}. The models were trained only with real sequences. Compared with DL3+(MobileNet), MFFA-DL3+(MobileNet-p8) achieved superior performance of 2.5$\%$$\sim$3.0$\%$ better mDice and 3.3$\%$$\sim$3.8$\%$ better mIoU with less inference time.

\begin{table}[!t]
\caption{\vspace{-0.3em} Segmentation Performance on ROBUST-MIS-Proctocolectomy}
\label{Tab_endovis_seg}
\vspace{-1.em}
\centering
\begin{tabular}{c|c|c|c}
\hline
\multirow{2}*{Model(Backbone)} & mDSC & mIoU & Time\\
& (\%) & (\%) & (ms)\\
\hline
\hline
TernausNet-16 \cite{iglovikov2018ternausnet} (MobileNet) & 79.6(2.9) & 71.6(3.0) & 21.6\\
LWANet \cite{ni2019attention} (MobileNet) & 76.2(3.2) & 67.6(3.3) & 14.2\\
MAFA-DL3+ \cite{qin2020towards} (MobileNet) & 81.7(2.8) & 74.1(3.0) & 27.8\\
\hline
DL3+ \cite{chen2018encoder} (MobileNet) & 78.1(3.1) & 69.9(3.2) & 9.0\\
DL3+ \cite{chen2018encoder} (MobileNet-p8) & 78.4(3.8) & 70.4(4.0) & 6.4\\
MFFA-DL3+ \cite{chen2018encoder} (MobileNet-p8) & \textbf{81.0(3.1)} & \textbf{73.5(3.2)} & 8.5\\
\hline
\multicolumn{4}{l}{* The bold font indicates the best performance in the column of the}\\
\multicolumn{4}{l}{$~~$second group (last three rows).}
\end{tabular}
\vspace{-2.0em}
\end{table}

%

\section{Discussion}\label{discussion}

We propose MFFA that aggregates features both temporally and spatially for better segmentation. 
By distributing the feature extraction burden to each frame in the sequence, our method allows using lightweight encoders to reduce computation costs.

Fig. \ref{trans_rst_ex} shows some segmentation results. As shown in Fig. \ref{trans_rst_ex}(a-d), MFFA could help improve segmentation by reducing false negatives under many challenging conditions including reflection and blood compared to the baseline experiments. Moreover, we found that MFFA could help reduce the false positives on objects that are not parts of the human body such as gauze, as shown in Fig. \ref{trans_rst_ex}(e). 
In this work, we focus on semantic segmentation in which pixels are classified into different classes without separating different objects. When analyzing surgical videos that have several instruments at the same time, instance segmentation that treats different objects of the same class separately is more appealing as it provides more information regarding the surgical workflow. We will explore leveraging temporal information for instance segmentation as future work. 


Table \ref{Tab_seg} shows that the synthetic sequences helped achieve better segmentation performance than when only the real frame sequences were used for training, but Table \ref{Tab_ttest} shows that this superiority is still not significant enough with p-values of 0.51 and 0.13. 
There is still space left for improvement because the current synthesis method is relatively simple. 
We chose this method because
i) the sequences consist of only 4 frames with small instrument movements, so this method already provides acceptable approximations to the real sequences. 
ii) The statistical or other knowledge of instrument moving behaviors are not available in the studied datasets. The true instrument movement pattern could be considered in the future to generate more realistic sequences. One potential option is performing interpolation between two neighboring labeled frames to generate better synthetic trajectories. 
On the other hand, the generative adversarial networks should be explored for generating more realistic frames in the synthetic sequences \cite{marzullo2021towards}. 


\section{Conclusions}

In this work, we develop and validate a MFFA module that performs feature aggregation based on temporal and spatial relationships between frame pixels to improve instrument segmentation. By using MFFA, we can reduce the deep encoder to its trimmed version and decrease the computation costs. Another advantage of MFFA is that it can be easily combined with any segmentation model that has an encoder-decoder architecture. Also, we propose a simple but effective method that generates synthetic frame sequences to assist network training and compensate for the lack of densely labeled real frame sequences. In the future, we will further improve the sequence synthesis approach and apply the proposed method to surgical video instance segmentation.



%

\ifCLASSOPTIONcaptionsoff
  \newpage
\fi



\bibliographystyle{IEEEtran}
\bibliography{IEEEabrv,IEEEexample}
%



\end{document}